\documentclass[%
preprint,
 amsmath,amssymb,
 aps,
]{revtex4-2}[aps]

\usepackage{graphicx}
\usepackage{dcolumn}
\usepackage{bm}
\usepackage{comment}
\usepackage{float} 

\usepackage{xcolor}

\usepackage[normalem]{ulem}

\begin{document}

\title{On instabilities in neural network-based physics simulators}

\author{Daniel Floryan}\email{dfloryan@uh.edu}

\affiliation{
  Department of Mechanical Engineering, University of Houston, Houston, TX 77204, USA
}%

\date{\today}

\begin{abstract}
When neural networks are trained from data to simulate the dynamics of physical systems, they encounter a persistent challenge: the long-time dynamics they produce are often unphysical or unstable. We analyze the origin of such instabilities when learning linear dynamical systems, focusing on the training dynamics. We make several analytical findings which empirical observations suggest extend to nonlinear dynamical systems. First, the rate of convergence of the training dynamics is uneven and depends on the distribution of energy in the data. As a special case, the dynamics in directions where the data have no energy cannot be learned. Second, in the unlearnable directions, the dynamics produced by the neural network depend on the weight initialization, and common weight initialization schemes can produce unstable dynamics. Third, injecting synthetic noise into the data during training adds damping to the training dynamics and can stabilize the learned simulator, though doing so undesirably biases the learned dynamics. For each contributor to instability, we suggest mitigative strategies. We also highlight important differences between learning discrete-time and continuous-time dynamics, and discuss extensions to nonlinear systems. 
\end{abstract}

\maketitle

\section{Introduction}
\label{sec:intro}

The ability to accurately and efficiently predict the behavior of a dynamical system is fundamental to science and engineering. 
As data have become increasingly available, machine learning has been leveraged to learn models that can accurately and efficiently simulate the dynamics of physical systems, both in cases when the true dynamics are unknown, or when they are known but are expensive to compute \cite{pathak2018model, vlachas2018data, linot2020deep, karniadakis2021physics, pfaff2021learning, linot2022data, srinivasan2022parallel, vlachas2022multiscale, chen2022automated, floryan2022data}. Weather forecasting offers a prime example of this new approach, where decades of high-quality data \cite{hersbach2020era5} have been used to develop deep learning-based forecasters with accuracy comparable to traditional numerical weather prediction while offering significant speedups \cite{pathak2022fourcastnet, bi2023accurate, lam2023learning}. 

While accurate at short times, neural network-based physics simulators can produce unphysical or unstable predictions at long times \cite{vlachas2018data, pfaff2021learning, linot2022data, stachenfeld2022learned, keisler2022forecasting, chattopadhyay2023long}. Although certain stabilization tricks have been identified, it is unclear when they may be employed and why they succeed. Here, we identify probable culprits for the origin of these instabilities by analyzing the training dynamics of neural networks tasked with emulating a dynamical system. Along the way, we propose mitigative strategies and rationalize stabilization tricks found in the literature.

To render the problem analytically tractable, we consider linear dynamical systems, which can be perfectly emulated by a linear single-layer neural network. Our analysis considers a prototypical neural network training scheme, consisting of the mean squared error loss function, Glorot weight initialization \cite{glorot2010understanding}, and gradient descent to update the weights of the neural network. We analyze discrete- and continuous-time dynamical systems in turn, highlighting important differences that arise between these two model classes. Our analysis focuses on the effects of energy distribution, weight initialization, and noise on the training dynamics. 

\section{Discrete-time dynamical systems}
\label{sec:disc}

Consider the linear discrete-time dynamical system
\begin{equation}
  x_{i+1} = Ax_i, \qquad x_i \in \mathbb{R}^n, A \in \mathbb{R}^{n \times n}.
\end{equation}
Suppose that $A$ is not known explicitly, and we want to learn an accurate approximation $\hat A \in \mathbb{R}^{n \times n}$ from a dataset of pairs of snapshots, $\{(x_i, x_i^\#)\}_{i=1}^m$, that are generated by the dynamical system, so that
\begin{equation}
  x_i^\# = Ax_i, \qquad i = 1, \ldots, m.
\end{equation}

Assembling the data into the data matrices $X = [x_1 \; \cdots \; x_m]$ and $X^\# = [x_1^\# \; \cdots \; x_m^\#]$, which are related by $X^\# = AX$, we seek the $\hat A$ that minimizes the mean squared error loss function $L$,
\begin{equation}
  \label{eq:lin1}
  \min_{\hat A} \quad \frac{1}{2mn} \Vert X^\# - \hat AX \Vert_F^2.
\end{equation}
Here, $\Vert \cdot \Vert_F$ is the Frobenius norm, the factor of $\frac{1}{2}$ is for convenience, and the factor of $\frac{1}{mn}$ mirrors how this loss function is typically implemented in machine learning software packages. The minimizer may not be unique. The least-squares/minimum-norm solution is $\hat A = X^\# X^+$, where $X^+$ is the pseudoinverse of $X$. However, what is of interest here is what the commonly used gradient descent algorithm produces. 

The gradient of $L$ with respect to $\hat A$ is
\begin{equation}
  \nabla_{\hat A}L = -\frac{1}{mn}(A - \hat A)XX^T.
\end{equation}
In the limit of infinitesimal learning rate, gradient descent leads to continuous-time gradient flow dynamics
\begin{equation}
  \frac{\textrm{d}}{\textrm{d}\tau} \hat A = -\nabla_{\hat A}L,
\end{equation}
with $\tau$ being a pseudo-time variable. The solution to this matrix ordinary differential equation is
\begin{equation}
  \hat A(\tau) = A + [\hat A(0) - A] \exp\left(-\frac{1}{mn}XX^T \tau \right),
\end{equation}
where $\hat A(0)$ is the initialization of our linear single-layer neural network. 

A coordinate change provides insight. Let $X = U\Sigma V^T$ be the full singular value decomposition of $X$, and $\tilde{A} = U^T AU$ the dynamics matrix in the basis of the left singular vectors. Then the training dynamics give
\begin{equation}
  \hat{\tilde{A}}(\tau) = \tilde{A} + [\hat{\tilde{A}}(0) - \tilde{A}]\exp\left(-\frac{1}{mn}\Sigma\Sigma^T \tau \right).
\end{equation}
Three points of interest arise. First, columns of $\hat{\tilde{A}}$ corresponding to non-zero singular values converge to the corresponding columns of $\tilde{A}$---that is, the true dynamics---while those corresponding to zero singular values remain equal to their initial values. In other words, the dynamics are not learnable in directions in which the data have no energy. Second, the rate of convergence depends on how energy is distributed in the data. In directions in which the data have low energy, convergence to the true dynamics is slower. It is, therefore, more difficult to learn the dynamics of low-energy modes. The convergence rates can be made uniform by normalizing the data so that they have equal energy in all directions. Third, the initialization of the neural network will impact the learned dynamics in the unlearnable directions, or in all directions if gradient descent is stopped short of convergence (which is typical). 

\begin{figure}
  \begin{center}
  \includegraphics[width=\linewidth]{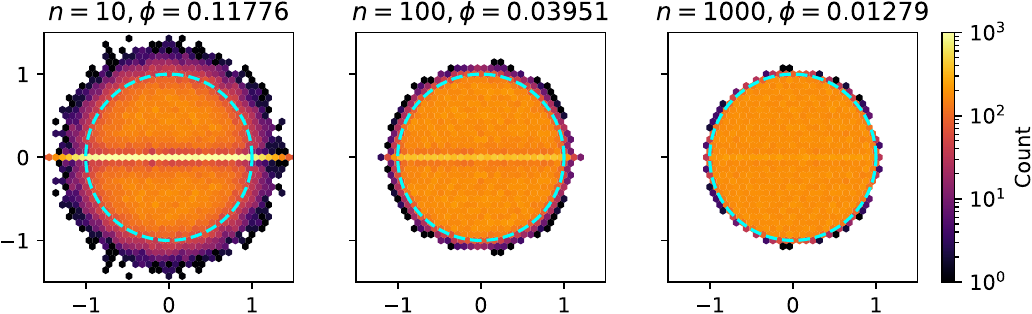}
  \end{center}
  \caption{Histogram of eigenvalues of an $n \times n$ matrix whose entries are generated using the Glorot normal initializer. $10^5/n$ realizations of the random matrix were used. The unit circle is drawn with a dashed cyan line. $\phi$ gives the fraction of eigenvalues outside of the unit circle. The Glorot uniform initializer produces nearly identical histograms. }
  \label{fig:randNormEvals}
\end{figure}

To illuminate the impact of the initialization, we suppose the data are in a set that is invariant under the dynamics of our system. This is generally true for the large class of physical systems with dissipation, for which the long-time dynamics approach an invariant manifold \cite{hopf1948mathematical, foias1988inertial, temam1994estimates, doering1995applied}. In our case, the true dynamics take the form
\begin{equation}
  \tilde{A} = \begin{bmatrix} \tilde{A}_{11} & \tilde{A}_{12} \\ 0 & \tilde{A}_{22} \end{bmatrix},
\end{equation}
where the upper-left submatrix gives the dynamics in the invariant subspace containing the data, and the other columns correspond to the unlearnable directions. As $\tau \rightarrow \infty$, the learned dynamics will be
\begin{equation}
  \hat{\tilde{A}} = \begin{bmatrix} \tilde{A}_{11} & \hat{\tilde{A}}_{12}(0) \\ 0 & \hat{\tilde{A}}_{22}(0) \end{bmatrix}.
\end{equation}
The eigenvalues of the learned dynamics are the union of the eigenvalues of $\tilde{A}_{11}$ and the eigenvalues of $\hat{\tilde{A}}_{22}(0)$. If $X$ has rank $r$ and $\hat A(0)$ is initialized using the typical Glorot normal or uniform initializers, then as $n \rightarrow \infty$, the distribution of the eigenvalues of $\hat{\tilde{A}}_{22}(0)$ converges almost surely to the uniform distribution on the disk of radius $\sqrt{\frac{n-r}{n}}$ centered at the origin \cite{tao2010random}. For finite $n$, however, the eigenvalues can lie outside the unit circle, as shown in Figure~\ref{fig:randNormEvals}, leaving the potential for unstable dynamics. 

This point bears emphasizing. In physical systems with dissipation, the state of the system will quickly approach an invariant set. When we collect measurements of the state, those measurements will generally not include information outside of the invariant set due to its stability. Without such information, it is impossible to learn the dynamics outside of the invariant set; the learned dynamics in that region depend on the initialization. Ironically, the stability of the true dynamics can lead to unstable learned dynamics since there is no information available to erase the memory of the potentially unstable initialization. 

\begin{figure}
  \begin{center}
  \includegraphics[width=\linewidth]{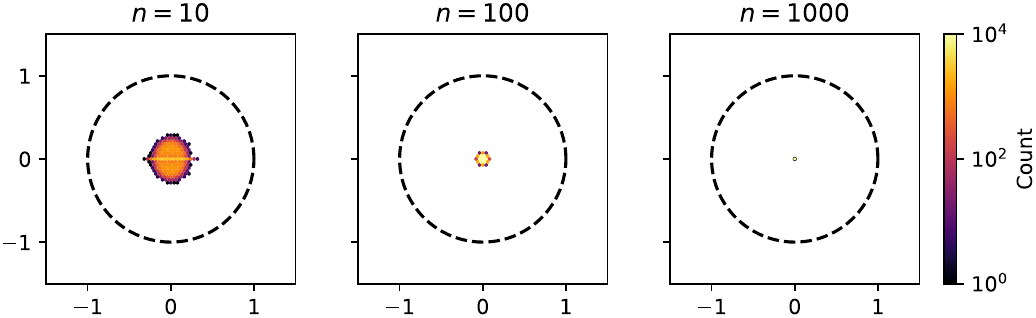}
  \end{center}
  \caption{Histogram of eigenvalues of an $n \times n$ matrix whose entries are generated using the initializer based on Gershgorin's circle theorem. $10^5/n$ realizations of the random matrix were used. The unit circle is drawn with a dashed black line. }
  \label{fig:gershgorin}
\end{figure}

Two remedies are clear. One is to restrict the learned model to remain in the invariant set. For linear systems, this amounts to projecting the system onto the subspace with non-zero singular values. The second remedy is to ensure that the initialization gives stable dynamics, which requires developing new weight initialization schemes. Gershgorin's circle theorem provides one way to create a mostly random matrix whose eigenvalues are guaranteed to lie inside the unit circle, thereby giving stable dynamics. For example, by drawing the entries of $\hat A(0)$ from the uniform distribution on $\frac{1}{n-1} [-1, 1]$ and setting the diagonal entries equal to zero, the eigenvalues of $\hat A(0)$ are bounded by the unit circle. This bound is sharp (e.g., if all off-diagonal entries are equal to $\frac{1}{n-1}$ then 1 is an eigenvalue) but almost surely conservative, as demonstrated in Figure~\ref{fig:gershgorin}. The distribution of eigenvalues can be expanded (but still bounded by the unit circle) by re-normalizing each row (or column) of $\hat A(0)$ so that the row (column) sums of the absolute values of the entries in each row (column) are equal to 1. 

\subsection{Noisy measurements}
\label{sec:noisy}

Curiously, it has been observed that injecting synthetic noise into the data during training can help stabilize learned physics simulators, though it is unclear why \cite{vlachas2018data, vlachas2020backpropagation, sanchez2020learning, pfaff2021learning, stachenfeld2022learned, su2022adversarial}.

Suppose our measurements are noisy, so that the data matrices are $X + N$ instead of $X$, and $X^\# + N^\#$ instead of $X^\#$. $N$ and $N^\#$ are random matrices of noise, with all entries assumed to be independent random variables with zero mean and variance $\sigma^2$, and independent of the noise-free data. It is likely that many columns of $N^\#$ are the same as columns of $N$, but shifted over one column in location, since data are often gathered from long trajectories that provide many data. With noisy data, the gradient of the loss function with respect to $\hat A$ is
\begin{equation}
  \nabla_{\hat A}L =  -\frac{1}{mn}[(A - \hat A)(X + N) + N^\# - AN](X + N)^T.
\end{equation}
In the basis of the left singular vectors of the noise-free data $X$, the training dynamics give
\begin{multline}
  \hat{\tilde{A}}(\tau) = \tilde{A} + [\hat{\tilde{A}}(0) - \tilde{A}] \exp\left( -\frac{1}{mn}(\Sigma V^T + U^T N)(\Sigma V^T + U^T N)^T \tau \right) \\
  + \frac{1}{mn}(U^T N^\# - \tilde{A}U^T N)(V\Sigma^T + N^T U) \left( -\frac{1}{mn}(\Sigma V^T + U^T N)(\Sigma V^T + U^T N)^T \right)^{-1}  \\
   \times \left[ \exp\left( -\frac{1}{mn}(\Sigma V^T + U^T N)(\Sigma V^T + U^T N)^T \tau \right) - I \right],
\end{multline}
where we have assumed that $(\Sigma V^T + U^T N)(\Sigma V^T + U^T N)^T$ is invertible, which is generally true for $m \ge n$. In the limit $\tau \rightarrow \infty$,
\begin{equation}
  \hat{\tilde{A}} = \tilde{A} + (U^T N^\# - \tilde{A}U^T N)(V\Sigma^T + N^T U) \left( (\Sigma V^T + U^T N)(\Sigma V^T + U^T N)^T \right)^{-1}. 
\end{equation}
The expected value is
\begin{equation}
  \hat{\tilde{A}} = \tilde{A}[I - m\sigma^2 ( \Sigma \Sigma^T + m\sigma^2 I )^{-1}],
\end{equation}
where we have used the fact that $\mathbb{E}[U^T NN^T U] = \mathbb{E}[NN^T] = m\sigma^2 I$. Rewriting the above expression provides clarity:
\begin{equation}
  \hat{\tilde{A}} = \tilde{A}
  \begin{bmatrix} \frac{\sigma_1^2}{\sigma_1^2 + m\sigma^2} & & & & \\
  & \ddots & & & \\
  & & \frac{\sigma_r^2}{\sigma_r^2 + m\sigma^2} & & \\
  & & & 0 & \\
  & & & & \ddots \end{bmatrix}.
\end{equation}

We see that when measurement noise is present, $\hat{\tilde{A}}$ is a biased version of $\tilde A$. In particular, columns of $\hat{\tilde{A}}$ corresponding to non-zero singular values converge to the corresponding columns of $\tilde A$, but biased by a multiplicative factor $\sigma_i^2/(\sigma_i^2 + m\sigma^2) \le 1$, where $\sigma_i$ is the $i^{\text{th}}$ singular value of $X$. This bias factor can be written as $SNR_i/(1 + SNR_i)$, where $SNR_i = \sigma_i^2/m\sigma^2$ is the signal-to-noise ratio in the direction of the $i^{\text{th}}$ singular vector. In contrast, without measurement noise, these columns of $\hat{\tilde{A}}$ converge to the true columns of $\tilde {A}$. The columns corresponding to zero singular values converge to zero, while they stayed equal to their initial guess when there was no measurement noise. The measurement noise adds damping to the training dynamics, erasing the memory of the initialization in the unlearnable directions and replacing it with dynamics that are strongly stable. This is a highly desirable effect since, as previously explained, in physical systems with dissipation, the stabilizing effect of dissipation is what makes those directions unlearnable. 

How fast is the memory of the initialization erased by the noise? The corresponding eigenvalues of the gradient flow dynamics are, in expectation, equal to $-\sigma^2/n$. The rate of convergence depends on the strength of the noise and the dimension of the system, with weak noise and a large dimension of the system leading to slow convergence. 

\begin{figure}
  \begin{center}
  \includegraphics[width=0.7\linewidth]{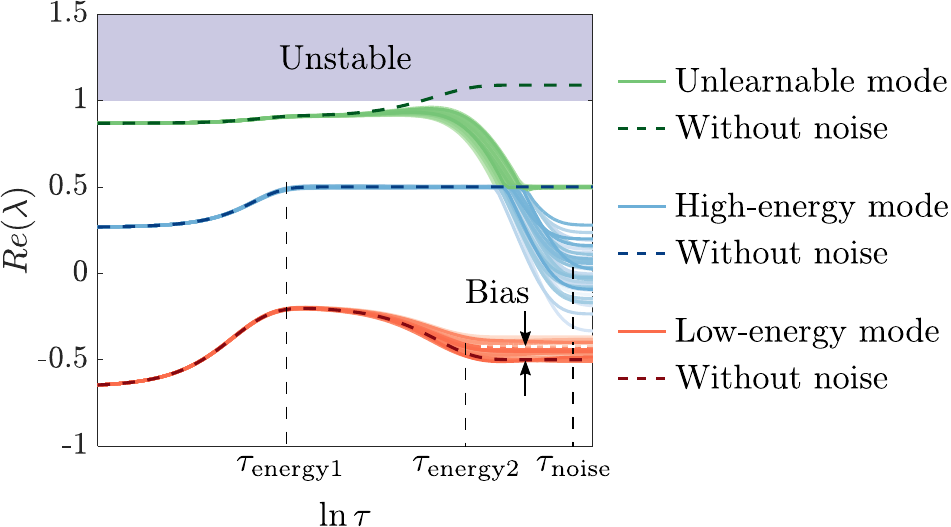}
  \end{center}
  \caption{Training dynamics of a three-dimensional system. The real parts of the eigenvalues are shown. The high-energy mode converges more quickly (blue; in time $\tau_{\text{energy1}}$) than the low-energy mode (red; in time $\tau_{\text{energy2}}$). Noise stabilizes the unlearnable and otherwise unstable mode (green), but biases the learnable dynamics. }
  \label{fig:schem2}
\end{figure}

There is a tradeoff between the desirable and undesirable effects of noise: noise stabilizes a learned physics simulator, with stronger noise stabilizing the learned system more quickly, but stronger noise also leads to greater bias. This is illustrated in Figure~\ref{fig:schem2} for a three-dimensional system, along with the effect of energy distribution. It may be possible to obtain stability while avoiding bias by selectively applying noise only in the unlearnable/zero-energy directions. 

\section{Continuous-time dynamical systems}
\label{sec:cont}

Important differences arise in continuous-time dynamics. Consider the linear continuous-time dynamical system
\begin{equation}
  \frac{\textrm{d}}{\textrm{d}t} x = Ax, \qquad x(t) \in \mathbb{R}^n, A \in \mathbb{R}^{n \times n}.
\end{equation}
As in the discrete-time case, suppose that $A$ is not known explicitly, and we want to learn an accurate approximation $\hat A \in \mathbb{R}^{n \times n}$ from a dataset of pairs of snapshots separated by a time $\Delta t$, $\{(x(t_i), x(t_i + \Delta t))\}_{i=1}^m = \{(x_i, x_i^\#)\}_{i=1}^m$, that are generated by the dynamical system, so that
\begin{equation}
  x_i^\# = e^{A\Delta t}x_i, \qquad i = 1, \ldots, m.
\end{equation}

Assembling the data into data matrices as before, we seek the $\hat A$ that minimizes the mean squared error $L$,
\begin{equation}
  \min_{\hat A} \quad \frac{1}{2mn} \Vert X^\# - e^{\hat A \Delta t}X \Vert_F^2. 
\end{equation}
Due to the presence of the matrix exponential, there is no simple expression for the gradient of $L$ with respect to $\hat A$, and the training dynamics are nonlinear in $\hat A$. To gain insight, we expand the matrix exponential to $\mathcal{O}(\Delta t)$---equivalent to solving the dynamics using the forward Euler method, making the learning problem equivalent to learning a residual network (scaled by $\Delta t$) \cite{he2016deep}. Under this approximation, the gradient of the loss function is
\begin{equation}
  \nabla_{\hat A}L = -\frac{\Delta t^2}{mn}(A - \hat A) XX^T,
\end{equation}
equal to that in the discrete-time setting but scaled by a factor $\Delta t^2$. The conclusions made in the discrete-time setting therefore extend to the continuous-time setting, but for the following exceptions. 

First, the factor of $\Delta t^2$ in the gradient changes the rate of convergence of the training dynamics. Small $\Delta t$ could lead to very slow convergence. 

Second, for continuous-time dynamics, an instability arises when the real part of any eigenvalue of $\hat A$ is positive. When $\hat A(0)$ is created using the Glorot initializer, each of its eigenvalues has a 50\% of having a positive real part. The continuous-time problem is therefore more susceptible to instability. Gershgorin's circle theorem can again be used to create a stable mostly random matrix by having the Gershgorin disks lie in the left half of the complex plane (or within the stable region of the numerical integrator begin used; for example, if using a residual network, the Gershgorin disks should be made to lie within the stability boundary of the forward Euler method). 

Finally, the effects of measurement noise differ in the details. With noisy data, the gradient is
\begin{equation}
  \nabla_{\hat A}L = - \frac{\Delta t^2}{mn}(A - \hat A)(X + N)(X + N)^T - \frac{\Delta t}{mn} (N^\# - N)(X + N)^T + \frac{\Delta t^2}{mn}AN(X + N)^T. 
\end{equation}
In the basis of the left singular vectors of the noise-free data $X$, the training dynamics give
\begin{multline}
  \hat{\tilde{A}}(\tau) = \tilde{A} + [\hat{\tilde{A}}(0) - \tilde{A}] \exp\left( -\frac{\Delta t^2}{mn}(\Sigma V^T + U^T N)(\Sigma V^T + U^T N)^T \tau \right) \\
  + \frac{\Delta t}{mn}[U^T N^\# - (I + \tilde{A}\Delta t)U^T N](V\Sigma^T + N^T U) \\
  \times  \left( -\frac{\Delta t^2}{mn}(\Sigma V^T + U^T N)(\Sigma V^T + U^T N)^T \right)^{-1}  \\
   \times \left[ \exp\left( -\frac{\Delta t^2}{mn}(\Sigma V^T + U^T N)(\Sigma V^T + U^T N)^T \tau \right) - I \right].
\end{multline}
In the limit $\tau \rightarrow \infty$,
\begin{equation}
  \hat{\tilde{A}} = \tilde{A} + \frac{1}{\Delta t}[U^T N^\# - (I + \tilde{A}\Delta t)U^T N](V\Sigma^T + N^T U) \left( (\Sigma V^T + U^T N)(\Sigma V^T + U^T N)^T \right)^{-1}. 
\end{equation}
The expected value is
\begin{equation}
  \hat{\tilde{A}} = \tilde{A}[I - m\sigma^2 ( \Sigma \Sigma^T + m\sigma^2 I )^{-1}] - \frac{m\sigma^2}{\Delta t}(\Sigma\Sigma^T + m\sigma^2 I)^{-1}.
\end{equation}Rewriting the above expression provides clarity:
\begin{equation}
  \hat{\tilde{A}} = \tilde{A}
  \begin{bmatrix} \frac{\sigma_1^2}{\sigma_1^2 + m\sigma^2} & & & & \\
  & \ddots & & & \\
  & & \frac{\sigma_r^2}{\sigma_r^2 + m\sigma^2} & & \\
  & & & 0 & \\
  & & & & \ddots \end{bmatrix}
   - \frac{1}{\Delta t}
   \begin{bmatrix} \frac{m\sigma^2}{\sigma_1^2 + m\sigma^2} & & & & \\
  & \ddots & & & \\
  & & \frac{m\sigma^2}{\sigma_r^2 + m\sigma^2} & & \\
  & & & 1 & \\
  & & & & \ddots \end{bmatrix}.
\end{equation}

As in the discrete-time setting, noise creates a multiplicative bias factor. Additionally, there is an additive bias that can be substantial for small $\Delta t$. The columns corresponding to zero singular values converge to columns whose only non-zero entries are along the diagonal and are equal to $-1/\Delta t$. Measurement noise again erases the memory of the initialization in the unlearnable directions and replaces it with stable dynamics, with smaller $\Delta t$ leading to more stable dynamics. In addition to the tradeoffs noted in the discrete-time setting, in the continuous-time setting $\Delta t$ also has a tradeoff: smaller $\Delta t$ creates more stable dynamics but greater bias.

\section{Discussion}
\label{sec:dis}

Despite the importance of the long-term stability of learned physics simulators, theoretical insight into this issue is conspicuously missing. Our findings for linear dynamical systems constitute an important step towards a general theory, buoyed by nonlinear analogs with strong empirical support. 

In the linear case, we showed that it is more difficult to learn low-energy dynamics due to the associated slower rates of convergence in the training dynamics. This seems to hold for nonlinear systems as well, for which it has been empirically observed that it is difficult to learn the dynamics of high wavenumbers in physical systems, which typically have low energy \cite{chattopadhyay2023long, lippe2024pde}. Although the difficulty to learn the dynamics of high wavenumbers has previously been attributed to the spectral bias of neural networks \cite{chattopadhyay2023long, xu2019frequency, rahaman2019spectral}, recent work on multi-stage neural networks supports that non-uniform distribution of energy and spectral bias both contribute to slow convergence \cite{wang2024multi}. The non-uniformity of energy could be addressed by first transforming the variables to a space in which the data are distributed isotropically, then learning the dynamics of the transformed variables. 

We then showed that dynamics off of the data subspace cannot be learned, and what is learned in the complement of the data subspace depends on the weight initialization. This obviously holds for nonlinear systems, where the dynamics off of the data submanifold cannot be learned. What is less clear is whether a non-trivial weight initialization scheme can be designed so that the default dynamics are stable, as we have done here for linear dynamical systems. In the linear case, an alternative is to project the system onto the data subspace. In the nonlinear case, manifold learning methods can be used to project the system onto the data submanifold; doing so has, indeed, been found to stabilize the learned dynamics \cite{linot2022data, chen2022automated, floryan2022data}. Another alternative is to add global damping to the system \cite{vlachas2018data, linot2022data, linot2023stabilized}, though it is unclear how strong it should be. 

Finally, we noted the empirical success of noise injection as a stabilizer when learning nonlinear dynamical systems, and showed why it works when learning linear dynamical systems. Adding noise to the data adds damping to the training dynamics. Damping is a generic mechanism that is likely to extend to the training dynamics of nonlinear systems. Furthermore, we showed that there is a tradeoff between the stabilization and bias that noise creates, and this tradeoff can be seen in empirical results for nonlinear systems \cite{sanchez2020learning}. The agreement with empirical results for nonlinear systems suggests that we have identified the correct mechanisms.

\bibliography{references}
\bibliographystyle{plain}

\end{document}